\documentclass{article}

\usepackage{arxiv}

\usepackage[utf8]{inputenc} 
\usepackage[T1]{fontenc}    
\usepackage{hyperref}       
\usepackage{url}            
\usepackage{booktabs}       
\usepackage{amsfonts}       
\usepackage{nicefrac}       
\usepackage{microtype}      
\usepackage{lipsum}
\usepackage{graphicx}
\graphicspath{ {./images/} }

\usepackage{cite}
\usepackage{amsmath,amssymb,amsfonts}
\usepackage{graphicx}
\usepackage{textcomp}
\usepackage{xcolor}
\usepackage{color}
\newcommand{\ignore}[1]{}
\usepackage{algorithm}
\usepackage[noend]{algorithmic}
\newtheorem{theorem}{Theorem}
\usepackage{multirow}

\usepackage{tikz}
\usetikzlibrary{arrows.meta, positioning}

\newcommand{\aneta}[1]{\textbf{\textcolor{blue}{Aneta: #1}}}

\author{
Thilina Pathirage Don\\
Optimisation and Logistics\\
School of Computer Science \\ and Information Technology\\
Adelaide University\\
Adelaide, Australia
\And
Aneta Neumann\\
Optimisation and Logistics\\
School of Computer Science \\ and Information Technology\\
Adelaide University\\
Adelaide, Australia
\And
Frank Neumann\\
Optimisation and Logistics\\
School of Computer Science \\ and Information Technology\\
Adelaide University\\
Adelaide, Australia
}

\title{Greedy Approaches for Packing While Travelling with Deterministic and Stochastic Constraints}

\begin{document}
\maketitle

\begin{abstract}
The travelling thief problem (TTP) is a well-known multi-component optimisation problem that captures the interdependence between two components: the tour across cities and the packing of items. The packing while travelling problem (PWT) is an NP-hard subproblem of TTP where the packing of items should be optimised for a given fixed tour. In many solvers, the packing component is often addressed using greedy heuristics. Here, the use of suitable greedy functions is essential for the success of greedy algorithms. In this paper, we introduce new reward functions tailored to the PWT and extend them to a hyper-heuristic framework to achieve further advantage. Furthermore, we investigate the chance constrained PWT for greedy approaches and adopt the newly introduced reward functions for stochastic weights. The experimental results clearly demonstrate the benefit of the tailored heuristics over the standard heuristics in both deterministic and stochastic constraints.
\end{abstract}




\section{Introduction}
\label{sec1}
Multi-component problems, which combine several problems in an interacting way, occur in many real-world applications. A prominent multi-component problem is the travelling thief problem (TTP), which has received significant interest in the evolutionary computation domain and heuristic search literature. Algorithms for TTP have to deal with the interactions between the two underlying problems, namely the travelling salesperson problem (TSP) and the knapsack problem (KP). Many solvers for TTP use packing heuristics for the KP, and they are often based on greedy approaches. Choosing the right greedy function is crucial for the success of greedy approaches. For classical problems such as TSP, neighbouring heuristics have been frequently applied, whereas for KP, sorting items according to profit-weight ratio is a popular approach. 

Greedy heuristics for TTP, such as the \textsc{Pack} algorithm~\cite{faulkner2015approximate}, have to take the information from the TSP and KP into account, and the question arises how the information from both problems should be combined into an appropriate greedy function. The goal of this paper is to investigate different greedy approaches for the packing part of TTP. Considering different tours, we examine the impact of using different reward functions for solving the packing problem, which is also known as the packing while travelling problem (PWT). Uncertainty in the problem components, such as the item weights and profits, can significantly affect aspects of a combinatorial problem, including fitness computation, reward calculation, and functionality of the constraints. It is interesting to see how the algorithms designed for the deterministic problem behave in the presence of stochastic elements and what level of adaptation they need to produce high-quality solutions. Hence, we further examine the impact of the greedy approaches under different constraint settings.

\subsection{Related Work}
\label{sec1_1}
Greedy algorithms are efficient for many classical polynomially solvable combinatorial optimisation problems, such as the computation of minimum spanning trees~\cite{kruskal1956shortest} and shortest paths~\cite{dijkstra1959note}. They have also been applied to many NP-hard problems and provide good approximations for problems such as vertex cover~\cite{gao2018randomized} and the optimisation of monotone submodular functions under different types of constraints~\cite{krause2014submodular}. However, they yield suboptimal solutions in such cases, and the choice of greedy functions is a key determinant for success. 

Since the introduction of the TTP~\cite{bonyadi2013travelling} and PWT~\cite{polyakovskiy2017packing,roostapour2019analysis}, many studies have investigated solutions to the classical problems and their variants.
Apart from the many studies focusing on a variety of evolutionary approaches~\cite{bonyadi2014socially,wu2017exact,yafrani2018efficiently,wuijts2019investigation,pathiragedon2025evolutionary}, the literature includes different heuristic approaches \cite{faulkner2015approximate,yafrani2016population,namazi2019profit,maity2020efficient,zhang2021solving} as well as studies on using hyper-heuristics~\cite{martins2017heuristic,yafrani2018hyperheuristic,ali2021hyperheuristic,rodriguez2022sequence,sarkar2024travelling}.
Chance constrained optimisation~\cite{miller1965chance} has been studied in relation to the KP~\cite{xie2019evolutionary,neumann2022evolutionary,li2024chance}. Solvers for KP have been introduced, and the insights obtained are useful for developing solvers for the TTP, as KP is embedded as a component in the TTP definition. For other optimisation problems such as submodular functions~\cite{doerr2020optimization,yan2023optimizing,neumann2025optimizing,neumann2021diversifying,yan2024sampling}, the vehicle routing problem~\cite{kohout1999time,bent2003dynamic,bent2007waiting}, the makespan scheduling problem~\cite{shi2022runtime} and the TTP~\cite{pathiragedon2024thechance,pathiragedon2025weighted}, have also been studied for the chance constraints.

\subsection{Our Contribution}
\label{sec1_2}
In this paper, we focus on a well-known greedy algorithm called \textsc{Pack}, originally introduced for the TTP. This algorithm introduces a reward function (also known as goodness value) that assigns a score to each item before the packing process begins. The original reward function is inspired by the profit-to-weight ratio commonly used to solve the KP greedily~\cite{neumann2022evolutionary} and incorporates new elements that account for the tour component.

To further improve item picking order, we introduce multiple reward functions for \textsc{Pack}, tailored to the exact benefit each item selection yields in the final solution. We then refine these reward functions to dynamically update the scores whenever an item is added to the packing plan. It takes into account the cost imposed by items already decided to collect on the items intended for pickup next. Finally, by using both existing and newly introduced reward functions, we develop a hyper-heuristic approach to gain further advantage.
To clearly visualise the advantages of the new reward functions for packing in experiments, we reduce the effect of the tour on the final solution by limiting it to fixed tours, a version of the problem known as PWT. We examine the behaviour of the greedy approaches in PWT with both deterministic and stochastic constraints.
To adopt reward functions for the stochastic environment, we first explore ways to represent stochastic weights using popular tail-bound inequalities. In total, we investigate the performance of four reward functions for the deterministic problem and two adaptations for stochastic variation. Furthermore, we test four algorithms for deterministic constraints and two for stochastic constraints, referring to the hyper-heuristic framework we proposed for packing. We observe a positive impact of tailored reward functions across all constraint settings for PWT, and the adaptability of these methods for the TTP.

The rest of the paper is as follows. Section~\ref{sec2} provides background on the classical PWT problem with the deterministic constraints and extends into the chance constrained variation with stochastic weights. The latter part of the section gives a brief introduction to the greedy heuristics found in TTP and the \textsc{Pack} algorithm. Section~\ref{sec3} describes the new tailored reward functions introduced for packing. In Section~\ref{sec4}, we present the adaptation of new reward functions to the PWT with stochastic constraints. Section~\ref{sec5} explains how the different heuristics are combined in a hyper-heuristic framework to gain further improvement. Section~\ref{sec6} provides details on the experimental setup and the results obtained. Finally, Section~\ref{sec7} gives the concluding remarks.

\section{Preliminaries}
\label{sec2}

The following sections explain the PWT under deterministic and stochastic constraints and then discuss the greedy heuristics applied in the \textsc{Pack} algorithm, which serve as the starting point for the novel methods introduced in this study.

\subsection{Packing While Travelling Problem}
\label{sec2_1}
 
The PWT has been formulated as a nonlinear knapsack problem based on the TTP. Similar to the definition of TTP, PWT involves a set of items of distinct profits and weights, distributed across a set of cities, skipping the first city. A vehicle (or an agent) starting from the first city, visits each city exactly once in a specific order, collecting items of its choice, and returns to the starting point. A collected item contributes to total profit but also slows the vehicle due to the added weight. A PWT solution includes two components: a tour and a packing plan. However, since the tour is predefined, the objective is to find the packing plan that maximises total benefit while satisfying the vehicle's capacity constraints. We refer to the formal definition of PWT as introduced in~\cite{polyakovskiy2017packing} in this section.

Given is a tour $x = ( x_1, \ldots, x_n ), x_i \in N$ as a sequence of a set of cities $N=\{1, \ldots, n\}$ with distances $d_{ij}$, $i, j \in N$, between them, and a set of items $M=\{e_1, \ldots, e_m\}$, where each item $e_i$ has a profit $p_i$ and a weight $w_i$, $1 \leq i \leq m$.
The vehicle visits all cities in the tour order $x$ and may pick up items along the way, provided that the total weight of the collected items does not exceed the vehicle capacity $B$.
The goal is to find a solution $s=(x,y)$ that consists of the given tour $x$ and a packing plan $y = ( y_1, \ldots, y_m) \in \{ 0,1 \}^m$ which maximises
\begin{equation}
    \label{alg: classical_pwt}
    z(x,y) = g(y) - R \left( \frac{d_{x_n x_1}}{\nu_{max}-\nu W_{x_n}} + \sum_{i=1}^{n-1} \frac{d_{x_i x_{i+1}}}{\nu_{max}-\nu W_{x_i}} \right)   
\end{equation}

\begin{equation}
    \label{classical_pwt - constraint}
    \text{subject to }  \sum_{j=1}^{m} w_j y_j \leq B.
\end{equation}

Here $g(y)=\sum_{j=1}^{m} p_{j} y_{j}$, which represents the total profit of the picked items. Equation~\ref{classical_pwt - constraint} provides a classical knapsack constraint, which states that the total weight of the picked items can not exceed the knapsack capacity $B$. The parameters $\nu_{\max}$ and $\nu_{\min}$ are the maximum and minimum traveling speed and $\nu = \frac{\nu_{max}-\nu_{min}}{B}$ is a normalising constant.
The accumulated weight of the items collected from city $x_1$ to city $x_i$ in tour $x$ is given by $W_{x_i}$. A cost factor (rental rate) $R$ is applied to the overall travel time.

\subsection{Stochastic Constraints for the PWT}
\label{sec2_2}

As described above, the classical PWT problem is a nonlinear knapsack problem that involves a deterministic weight constraint. To investigate the same problem under stochastic constraints, we assume that the weights assigned to each item are stochastic. With stochastic weights, we treat the knapsack constraint as a probabilistic constraint (a chance constraint), in which the probability of satisfying the knapsack bound should exceed a predefined threshold. Hence, the objective of the chance constrained PWT would be remodelled to find a solution that delivers the maximum benefit while meeting the knapsack constraint with a probability of at least $\alpha$, which is a larger value ($\alpha\geq 0.9$). 
We refer to the surrogate-based model~\cite{pathiragedon2024thechance} introduced for the chance constrained TTP, which involves stochastic weights, to formulate the PWT with stochastic constraints.

The surrogate-based model defines the chance constrained PWT as follows. The input is given by $n$ cities and $m$ items, where the stochastic weights of the items are random variables $\{w_1,...,w_m\}$. The uncertainty of the weights is represented by the expected values $\{\mu_1,...,\mu_m\}$ and the variances $\{\sigma^2_1,...,\sigma^2_m\}$. The profits $\{p_1,...,p_m\}$ of the items are deterministic. We aim to find a solution $s=(x,y)$ consisting of the predefined tour $x=\{x_1,...,x_n\}, x_i \in N$ and a packing plan $y=\{y_1,...,y_m\}, y_j \in \{ 0,1 \}$, that maximises the benefit, referring to the objective function of the PWT (\ref{alg: classical_pwt}),
\begin{equation}
    \label{chance constraint: for surrogate-based pwt}
    \text{subject to }  P_r\left(\sum_{j=1}^{m} w_j y_j \leq B \right) \geq \alpha.
\end{equation}

To estimate the probability that the constraints are satisfied, we refer to two popular tail inequalities, namely the Hoeffding bound (\ref{equ: hoeffding}) and Chebyshev's inequality (\ref{equ: chebyshev's inequality}). We consider each weight $w_i$ to be uniformly distributed in $[\mu_i - \delta, \mu_i + \delta]$ range, where the $\delta$ presents the uncertainty of the weights. It is important to note that other distributions can also be considered similarly. However, the applicability of tail inequalities depends on the underlying distributions. For independent and identically distributed random variables and a large value of $\alpha$, a Chernoff/Hoeffding-type surrogate function is recommended~\cite{neumann2022evolutionary}.
%
\begin{theorem}[Hoeffding bound]
    Let $y_1, \ldots, y_m$ be independent variables taking values in $[a_j, b_j]$ of length $c_j := b_j - a_j$. Let $y = \sum^n_{j=1}y_j$  Then for any $\lambda \geq 0$, 
    \begin{equation}
        \label{equ: hoeffding}
        \begin{aligned}
            & Pr[y \geq \mu + \lambda]\leq e^{-\frac{2\lambda^2}{\sum^n_{i=1}c_i^2}}
        \end{aligned}
    \end{equation}
\end{theorem}

Let $|y|_1$ be the number of items in the packing plan $y$. Based on the Hoeffding bound (\ref{equ: hoeffding}), the chance constraint is met if,
\begin{equation}
    \label{W_hat Hoeffing constraint}
    \hat{W}_{Hoe}(y) = \mu(y) + \delta \cdot \sqrt{2|y|_1 \cdot \ln\left(1/(1 - \alpha)\right) } \leq B.
\end{equation}

\begin{theorem}[Chebyshev's inequality]
    Let $y$ be a random variable with the expected value $\mu_y$ and the variance $var(y)$. Then, for any $\lambda \in \mathrm{R}^+ $, 
    \begin{equation}
    \label{equ: chebyshev's inequality}
    \begin{aligned}
    \Pr(y \geq \mu_y + \lambda) \leq \frac{var(y)}{var(y) + \lambda^2}.
    \end{aligned}
    \end{equation}
\end{theorem}

Let $w(y)= \sum_{j=1}^m w_jy_j$ be the weight of a given packing plan $y=(y_1,...,y_m)$, $\mu(y)= \sum_{j=1}^m \mu_jy_j$ be the expected weight,  and $var(y) = \sum_{j=1}^m \sigma_j^2 y_j$ be its variance.
Based on Chebyshev's inequality (\ref{equ: chebyshev's inequality}), the chance constraint is met if,
\begin{equation}
\label{W_hat Cheb constraint}
\hat{W}_{Cheb}(y) = \mu(y) + \sqrt{\alpha/(1-\alpha)}\cdot\sqrt{var(y)} \leq B. 
\end{equation}

\subsection{Packing Heuristics for TTP}
\label{sec2_3}

Here, we discuss the \textsc{Pack} algorithm~\cite{faulkner2015approximate}, introduced as a greedy solver for packing items in the deterministic TTP, and the heuristics incorporated, as our study is set in this context. In general, \textsc{Pack} refers to a reward function $r1$ to compute a score (goodness value) for each item and prioritise items for the packing based on the score. To generate the score $r1_{ik}$ for the item $k$ placed in city $i$, the algorithm refer to the given reward function  (\ref{reward-function: r1}), where $p_{ik}$ is the profit of the item, $w_{ik}$ is the weight of the item and $d_i$ is the distance from the city $i$ to the end of a given tour $x$.
\begin{equation}
    \label{reward-function: r1}
    r1_{ik} = \frac{p_{ik}^\gamma}{w_{ik}^\gamma \times d_i}   
\end{equation}

The reward function is further modified by introducing an exponent $\gamma$ for the weight and profit. The $\gamma$ value is useful when a parent algorithm, such as \textsc{PackIterative}~\cite{faulkner2015approximate}, iteratively executes the \textsc{Pack} with different exponents. In our study, we are interested in understanding how different types of reward functions influence the quality of a packing plan, and we consider the case $\gamma=1$ as a baseline for comparison.

Once the scores are assigned, the algorithm sorts the items in non-decreasing order and considers each item in turn, starting with the first. If the constraint is satisfied, the item is added to the packing plan $y$ and the algorithm calculates the objective score $Z$. If $Z$ is better than the current best objective score, $Z$ becomes the new best objective score, and $y$ becomes the new best packing plan. If not, the packing plan drops the added item and returns to the previous packing plan. When no further improvement is possible, the algorithm stops and returns the best packing plan. The combination of the tour and the final packing plan is the final solution.

As shown, the greedy algorithm primarily relies on the reward function to select items for the packing plan. Hence, it is important to design a tailored reward function to improve item selection. During our preliminary studies, we examined a range of reward functions with both simple and complex packing heuristics and narrowed the list based on the improvements they brought over existing methods. The next sections describe the new approaches we introduce to improve the function of rewarding items.

\section{Improved Heuristics: Tailoring the Packing Heuristics for PWT}
\label{sec3}

As explained in Section~\ref{sec2_1}, once an item is picked, it affects the total benefit of the solution in two different ways. First, it increases the total profit by adding the profit of the picked item. Second, the weight of the picked item increases the vehicle's total weight, slowing it down. An increase in profit increases total benefit, whereas an increase in travel cost has the opposite effect. So it would be beneficial to consider these two factors when deciding which item to pick. We propose a new reward approach $r2$ that better reflects this requirement in the context of PWT. In the $r2$ setting, we treat the reward of an item as its profit minus the travel cost difference incurred by including it in the packing plan. The gain from the cost is the difference between the cost of travelling without picking up the item and the cost of travelling with the item.

The rewards are calculated before packing begins, when no items have been added to the packing plan. If the item $k$ in city $i$ is not collected, we assume the vehicle travels a distance $d_i$ from the city $i$ to the end of the tour at maximum velocity $\nu_{max}$. If collected, the velocity is reduced due to the item's weight $w_{ik}$. The travel cost in each scenario can be computed by multiplying the travel time ($\frac{distance}{velocity}$) by the rental rate $R$. We use the cost difference between these two scenarios along with the profit of the item $p_{ik}$ to calculate the reward (\ref{reward-function: r2}).
\begin{equation}
    \label{reward-function: r2}
    r2_{ik} = p_{ik} - R \left(\frac{d_i}{\nu_{max} - \nu w_{ik}} - \frac{d_i}{\nu_{max}}\right) 
\end{equation}

Based on preliminary investigations, we figure that using the ratio between the $r2_{ik}$ and the weight of the item $w_{ik}$ also serves as an effective reward function (\ref{reward-function: r3}).
\begin{equation}
    \label{reward-function: r3}
    r3_{ik} = \frac{r2_{ik}}{w_{ik}}   
\end{equation}

The rewards are usually calculated once before packing begins, since the values of the elements used in the calculation are considered deterministic and will not change over time. However, it is clear that the travel costs would change whenever an item is added to the packing plan. Given the potential dynamic changes in travel costs, we modify $r2$ to an iterative reward function $r4$, which updates the rewards for all remaining items whenever an item is added to the packing plan. To re-calculate the reward for all the remaining items $\{e_1, e_3\}$, we refer to the $r4$ (\ref{reward-function: r2_t}). The function refers to the accumulated weight $W_i$ from the city $i$ to the end of the tour to calculate the current travel cost. If the item $k$ in city $i$ is selected, its weight $w_{ik}$ is added to $W_i$ to compute the updated travel cost. The cost difference will be incorporated in updating the item's reward.
\begin{equation}
    \label{reward-function: r2_t}
    r4_{ik} = p_{ik} - R \left(\frac{d_i}{\nu_{max} - \nu(W_i + w_{ik})} - \frac{d_i}{\nu_{max} - \nu W_i}\right) 
\end{equation}

Further, we take the ratio between $r4_{ik}$ and the weight of the item $w_{ik}$ also as a new reward function (\ref{reward-function: r3_t}).
\begin{equation}
    \label{reward-function: r3_t}
    r5_{ik} = \frac{r4_{ik}}{w_{ik}}   
\end{equation}

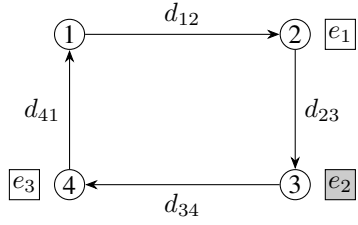
\begin{figure}[t]
    \centering
    \begin{tikzpicture}[
    node/.style={draw, circle, minimum size=4mm, inner sep=0pt},
    sq/.style={draw, rectangle, minimum size=4mm, inner sep=0pt},
    >=Stealth
]

\node[node] (n1) at (0,2) {1};
\node[node] (n2) at (3,2) {2};
\node[node] (n3) at (3,0)   {3};
\node[node] (n4) at (0,0)   {4};

\node[sq] (s2) at (3.6,2)  {$e_1$};
\node[sq, fill=gray!40] (s3) at (3.6,0)    {$e_2$};
\node[sq] (s4) at (-0.6,0)   {$e_3$};


\draw[->] (n1) -- node[above] {$d_{12}$} (n2);
\draw[->] (n2) -- node[right] {$d_{23}$} (n3);
\draw[->] (n3) -- node[below] {$d_{34}$} (n4);
\draw[->] (n4) -- node[left]  {$d_{41}$} (n1);

\end{tikzpicture}
    \caption{Example TSP tour to represent the weight accumulation in $r4$ reward function. See the discussion in the text.}
    \label{fig: pwt-diagram-01}
\end{figure}

The weight of an item in the packing plan may or may not affect the travel cost of a remaining item, depending on the position of the remaining item in the tour relative to the picked item. 
To explain how it works, we refer to the Figure~\ref{fig: pwt-diagram-01}, which represents a sample TSP tour $x^*=[1, 2, 3, 4, 1]$ in a setup with four cities $\{1, 2, 3, 4\}$ and three items $\{e_1, e_2, e_3\}$. The distances between adjacent cities $i$ and $j$ are denoted as $d_{ij}$. Assume the item $e_2$ is already in the packing plan.
The weight of the picked item $e_2$ will affect the cost of travelling from city $2$ to the end of the tour, as $e_2$ lies between them. Hence, the reward calculation for the remaining item $e_1$ will be affected by the weight of $e_2$. However, the picked item $e_2$ won't affect the cost of travelling from city $4$ to the end of the tour. Hence, the reward calculation of item $e_3$ does not consider the weight of $e_2$.

To reflect these iterative reward calculations in the \textsc{Pack} algorithm, we introduce a modified version as presented in Algorithm~\ref{alg: Pack-IH} ($\textsc{Pack}_{IH}$). Once an item is added to the packing plan, the algorithm removes the selected item from the list of items $M$, updates the scores based on new accumulated weights and sorts the list in non-decreasing order. Then the counter $c$ is set to zero to begin at the top of the reordered list of items. If the item is not picked, a recalculation of rewards is not needed. The algorithm restores the current best packing plan and increments the counter $c$ to refer to the next item in the ordered list. This version does not include the technique referred to in the original \textsc{Pack}~\cite{faulkner2015approximate} to reduce the number of fitness calculations, as it is important to capture the gain from adding each item. The rest of the algorithm works similarly to the classical counterpart.
\ignore{\aneta{It would be great ti improve the presentation of the Algorithm 1 and 2.} fixed.}

\begin{algorithm}[t]
\caption{$\textsc{Pack}_{IH}(x^*)$}
\begin{algorithmic}[1]
\STATE Compute a score for each of the items $I_m \in M$
\STATE Sort the items in non-decreasing order of their scores
\STATE Set the current packing plan $P = \emptyset$ and total weight $W' = 0$
\STATE Set the best packing plan $P^* = \emptyset$
\STATE Set the best objective value $Z^* = -\infty$
\STATE Set the counter $c=0$
\WHILE{$(W' \leq B)$ and $(k \leq |M|)$}
\IF{$(W'+ w_k \leq B)$}
    \STATE Add item $I_k \in M$ to the packing plan $P=P\cup{I_k}$
    \STATE Set $W'=W' + w_k$
    \STATE Compute the objective value $Z=Z(x,y)$
        \IF{$(Z<Z^*)$}
            \STATE Update $W' = W'- w_k$
            \STATE Restore the packing plan $P=P^*$
            \STATE Set $c=c+1$
        \ELSE
            \STATE Update the best packing plan with $P^*=P$
            \STATE Set $Z^*=Z$
            \STATE Set $M'=M$
            \STATE Remove selected item from the list $M = M' \setminus I_k$
            \STATE Update the score for each of items $I_m \in M$
            \STATE Sort the items in non-decreasing order of their scores
            \STATE Set $c=0$
        \ENDIF
\ENDIF
\ENDWHILE
\RETURN $P^*$
\end{algorithmic}
    \label{alg: Pack-IH}
\end{algorithm}

\section{Adaptation of Heuristic Algorithms for the Chance Constrained PWT}
\label{sec4}

As discussed earlier, the $\textsc{Pack}$ algorithm computes a reward for each item, which involves weights, sorts the items by score, and considers items for the packing plan from the sorted list. There are adaptations of the $\textsc{Pack}$ algorithm that can be directly applied in this context, such as  $\textsc{Pack}_{SF}$, $\textsc{Pack}_{S}$ and $\textsc{Pack}_{WS}$, which were originally developed for the chance constrained TTP~\cite{pathiragedon2024thechance,pathiragedon2025weighted}.
Since the definition of the chance constrained PWT introduced in Section~\ref{sec2_2} aligns with the surrogate functions, we refer to $\textsc{Pack}_{SF}$\cite{pathiragedon2024thechance} as a foundation for further improvements.
These adaptations refer to either the deterministic or the averaged weights to calculate rewards. In all of these approaches, the effect of stochasticity on the reward function is relatively small.

Hence, to adopt the aforementioned greedy approaches for the chance constrained PWT with stochastic weights, we first introduce a method to compute stochastic weights referring to the tail bound inequalities. Then, the new weight representation is applied to the reward functions introduced in Section~\ref{sec3} to produce a set of optimised reward functions for the item selection in a chance constrained setting.
In our approach, we apply a similar technique employed by Neumann et al.~\cite{neumann2022evolutionary} to compute the increased expected values for the weights ($w'$) in the reward calculation. Both the Hoeffding bound and Chebyshev's inequality are used to estimate the uncertainty.

First, we refer to the Hoeffding bound to obtain an increased expected weight $w'_{ik}$ for the item $k$ placed in city $i$. The increased expected weight ($w'_{ik}$) modifies the expected weight ($\mu_{ik}$) by adding uncertainty ($u$). The uncertainty is based on the packing plan $y$ and the impact if item $k$ is added to $y$ (Equation~\ref{alg: increased-expected-weight}). It is important to note that the increased expected weight should be calculated each time an item is added to the packing plan.
\begin{equation}
\label{alg: increased-expected-weight}
w'_{ik} = \mu_{ik} + u(y,k)
\end{equation}

Referring to the Hoeffding bound, we derive the following to estimate the uncertainty, where $|y|_1$ is the number of items in the packing plan. Since the packing plan initially contains no items, the algorithm refers to Equation~\ref{alg: increased-expected-weight-for-the-first-round} to generate the increased expected weights.
\begin{equation}
    \label{alg: increased-expected-weight-for-the-first-round}
     u(y,k) = \delta_i \cdot \sqrt{2 \cdot \ln\left(1/(1 - \alpha)\right)}
\end{equation}

The algorithm uses $w'_{ik}$ in the reward function. As usual, the algorithm then sorts all items, selects one item at a time, and considers it for the packing plan. After the first item is added to the packing plan, the algorithm updates the $w'_{ik}$ of all remaining items, referring to the Equation~\ref{alg: increased-expected-weight-detailed} and recalculates the goodness values. Then it picks the first item on the sorted list and considers it for the packing plan.

\begin{equation}
    \label{alg: increased-expected-weight-detailed}
    u(y,k) = \delta_i \cdot \left( \sqrt{2(|y|_1+1) \cdot \ln\left(1/(1 - \alpha)\right)} - \sqrt{2|y|_1 \cdot \ln\left(1/(1 - \alpha)\right)} \right)
\end{equation}

Similarly, we refer to Chebyshev's inequality and compute the increased expected weights. Since the packing plan initially contains no items, the algorithm refers to the Equation~\ref{alg: increased-expected-weight-for-the-first-round-Cheb}, which generates the increased expected weights and uses them to compute the rewards for items.

\begin{equation}
    \label{alg: increased-expected-weight-for-the-first-round-Cheb}
    w'_{ik} = \mu(w_{ik}) + \sqrt{\alpha/(1-\alpha)}\cdot\sqrt{var(y)} 
\end{equation}

Subsequently, the algorithm sorts all items, selects one item at a time, and considers it for the packing plan. Every time an item is added to the packing plan, the algorithm updates the $w'_{ik}$ of all remaining items, referring to the Equation~\ref{alg: increased-expected-weight-detailed-Cheb}.

\begin{equation}
    \label{alg: increased-expected-weight-detailed-Cheb}
    w'_{ik} = \mu(w_{ik}) + \sqrt{\alpha/(1-\alpha)}\cdot \left( \sqrt{var(y')} - \sqrt{var(y)} \right)
\end{equation}

The variance of the new packing plan $y'$ can be calculated as $var(y') = var(y) + \sigma_k^2$, where the $\sigma_k^2$ is the variance of the item $k$ in city $i$. Then, the algorithm recalculates the goodness value, sorts the item list, and selects the first item for consideration in the packing plan.
To use this new weight representation in iterative reward functions, including $r4$ (\ref{reward-function: r2_t}) and $r5$ (\ref{reward-function: r3_t}), we simply swap the deterministic weight $w$ with the stochastic weight $w'$, to produce the reward functions $r6$ (\ref{reward-function: r6}) and $r7$ (\ref{reward-function: r7}) for the PWT with the stochastic constraints.

\begin{equation}
    \label{reward-function: r6}
    r6_{ik} = p_{ik} - R \left(\frac{d_i}{\nu_{max} - \nu(W_i + w'_{ik})} - \frac{d_i}{\nu_{max} - \nu W_i}\right) 
\end{equation}

\begin{equation}
    \label{reward-function: r7}
    r7_{ik} = \frac{r6_{ik}}{w'_{ik}}   
\end{equation}

\section{Hyper-heuristics: Combined Use of Tailored Heuristics}
\label{sec5}

Hyper-heuristics serve as search methodologies to automate the heuristic design process by selecting low-level heuristics (LLHs) or by generating new heuristics that combine the components of LLHs~\cite{ross2003learning,yafrani2018hyperheuristic}.
We develop a hyper-heuristic framework (Algorithm~\ref{alg: HH-Framework}) that takes multiple reward functions as LLHs and iteratively improves the selection of heuristics for each item.

The process is as follows. First, the algorithm takes a list of reward functions ($H$) as inputs and uses them to initialise a combination of heuristics $hh$ of size $n$, where $n$ equals the total number of items. The $hh$ represents the order of the reward functions to be used for each item when picking.
Then the algorithm uses a modified version of \textsc{Pack} named $\textsc{Pack}_{HH}$ to take in the fixed tour $x^*$ and the $hh$ to generate the initial packing plan $y$. In order to do that, initially, $\textsc{Pack}_{HH}$ refers to the first LLH (reward function) in $hh$ and uses it to score items and orders based on the score. Once an item is picked, the algorithm removes it from the list, points the next LLH to recalculate the scores of the remaining items, and reorders them based on the new scores. The process continues until it considers all items for the packing plan and outputs the generated packing plan.

As the next step, the algorithm randomly flips each LLH in $hh$ to one of the other heuristics from $H$, with a given mutation rate and generates a new combination of heuristics $hh'$. Then, it uses $hh'$ to generate a new packing plan $y'$. The algorithm compares the two solutions to select the best combination of heuristics with the corresponding solution to carry forward to the next iteration. The process stops once the given stopping criteria are satisfied.
\ignore{\aneta{changed for clarification.} included the changes.}
We use two methods to initialise the heuristic combination $hh$. The simplest technique we use is to initialise $hh$ using the original reward function $r1$. The other technique is to run \textsc{Pack} with each reward function separately and pick the one that yields the best objective score to initialise $hh$. Based on the initialisation technique and the number of heuristics used, we identify six variations of the hyper-heuristic approach for PWT with both deterministic and stochastic constraints (Table~\ref{table: HH-variations}).

\begin{algorithm}[t]
\caption{\textsc{Hyper-heuristics Framework}}
\begin{algorithmic}[1]
    \STATE Input the packing heuristics $h_i \in H$ and fixed tour $x^*$.
    \STATE Initialise the heuristics combination $hh$ with size $n$.
    \STATE Generate the initial packing plan $y=\textsc{Pack}_{HH}$($x^*, hh$).
    \STATE Set the initial PWT solution $s=(x^*,y)$.
    \STATE Set mutation rate $\omega$
    \WHILE{stopping criterion is not satisfied}
        \STATE Generate $hh'$=mutate ($hh$, $\omega$).
        \STATE Generate the new packing plan $y'=\textsc{Pack}_{HH}$($x^*, hh'$).
        \STATE Set the new PWT solution $s'=(x^*,y')$.
        \IF{$f(s') \geq f(s)$}
            \STATE Set $hh \gets hh'$
            \STATE Set $y \gets y'$
            \STATE Set $s \gets s'$
        \ENDIF
    \ENDWHILE
    \RETURN $s, hh$
\end{algorithmic}
    \label{alg: HH-Framework}
\end{algorithm}

\begin{table}[t]
    \caption{Initialisation techniques and the reward functions used in the hyper-heuristic algorithms.}
    \label{table: HH-variations}
    \centering
\begin{tabular}{|l|l|l|l|l|}
    \hline
    problem type & algorithm & initialisation & heuristics used \\
    \hline
        deterministic & $HH_{1}$ & $r1$ & \{$r1, r2, r3$\} \\
        ~ & $HH_{2}$ & best & \{$r1, r2, r3$\} \\
        ~ & $HH_{3}$ & $r1$ & \{$r1, r2, r3, r4, r5$\} \\
        ~ & $HH_{4}$ & best & \{$r1, r2, r3, r4, r5$\} \\
        stochastic & $HH_{5}$ & $r1$ & \{$r1, r6, r7$\} \\
        ~ & $HH_{6}$ & best & \{$r1, r6, r7$\} \\
    \hline
\end{tabular}
\end{table}

\section{Experimental Investigations}
\label{sec6}
We conducted experiments to evaluate the performance of these greedy approaches for PWT under both deterministic and stochastic constraints. We test each algorithm on a range of benchmark TTP instances and analyse the results to gain insights.
\ignore{\aneta{extensive use of the word use} fixed.}
\ignore{\aneta{what is the justification for parameter $\delta$ and $\alpha$, mutation rate?} fixed.}

\subsection{Experimental Setup}
\label{sec6_1}
We refer to a set of comprehensive benchmark instances for TTP~\cite{polyakovskiy2014comprehensive} to construct PWT instances for our experiments. Each TTP instance includes a TSP component mapped with a KP component. The TSP component consists of a list of cities, each with coordinates in a Cartesian plane. The KP component lists the items, each with its profit, weight, and assigned city. The knapsack capacity $B$ and a renting rate $R$ are predefined in each instance. The number of cities in the instances ranges from $51$ to $1000$, with one item per city except for the first city. Two types of TTP instances are considered, where the weights and profits of an item are either strongly correlated (bsc) or uncorrelated (unc).
Unlike a TTP instance, a PWT instance includes a fixed tour. Hence, to construct the PWT instances, we first utilise the Chained Lin-Kernighan heuristics (CLK)~\cite{applegate2003chained} to generate $30$ random tours for each TTP instance, referring to its TSP component. Then, we couple each tour with the KP component from the same TTP instance to set up PWT instances. Altogether, we generate $300$ PWT instances by producing $30$ PWT instances for each of the $10$ TTP instances. The generated instances facilitate the evaluation of reward functions and hyper-heuristic algorithms.

For the experiments with stochastic constraints, we consider fixed uncertainties in the weights. We set a small uncertainty ($\delta=20$) and refer to the same $\delta$ value in all the benchmark instances to define uncertainty in weights. To evaluate the chance constraint, we set two different confidence levels ($\alpha$), where $\alpha \in \{ 0.9, 0.999 \}$. We set up $\alpha$ values in a way to utilise both tail inequalities introduced in Section~\ref{sec2_2}. The algorithms refer to Chebyshev's inequality for the smaller $\alpha$ and to the Hoeffding bound for the larger $\alpha$.
The stopping criterion for the hyper-heuristics algorithms is limited to $1000$ iterations due to their resource intensity. Based on preliminary studies, we associate a higher mutation rate of $0.1$ with the hyper-heuristic framework (Algorithm~\ref{alg: HH-Framework}) to gain sufficient benefit. In all experimental settings, we run each experiment independently with the $30$ PWT instances and compute the mean and standard deviation of the resulting objective scores.

We adopt \textsc{Pack} with certain modifications, in order to accommodate various heuristic applications in different constraint settings (see Table~\ref{table: Pack-variations}). For the deterministic constraints, we employ \textsc{Pack} as the base algorithm. The reward functions $r2$ and $r3$ are implemented in \textsc{Pack} by simply replacing the original reward function $r1$. To include the iterative reward functions ($r4$ and $r5$), we updated the base algorithm with a module that recalculates the scores, as in $\textsc{Pack}_{IH}$. For the simple hyper heuristic approaches ($HH_1$ and $HH_2$), we modify the base algorithm to take a combination of heuristics $hh$ as an additional input parameter and to do the selection of items based on $hh$, which is described as $\textsc{Pack}_{HH}$ in Section~\ref{sec5}. To handle hyper-heuristics that involve iterative reward functions ($HH_3$ and $HH_4$), we update $\textsc{Pack}_{HH}$ with an additional module to recalculate the scores, similar to $\textsc{Pack}_{IH}$. For the stochastic constraints, we use $\textsc{Pack}_{SF}$ as the base algorithm and make modifications whenever it accommodates iterative reward functions and hyper-heuristics in a similar way. It is important to note that none of the modifications affects the greedy algorithm's core functionality.
\begin{table}[t] 
    \caption{Modifications done for the \textsc{Pack} algorithm to implement the reward functions and algorithms.}
    \label{table: Pack-variations}
    \centering
\begin{tabular}{|l|l|l|}
    \hline
    heuristics/algorithm & $\textsc{Pack}$ modification \\
    \hline
        $r1, r2, r3$ & $\textsc{Pack}$ \\
        $r4, r5$ & $\textsc{Pack}$ for iterative reward functions \\
        $r6, r7$ & $\textsc{Pack}_{SF}$ for iterative reward functions \\
        $HH_1, HH_2$ & $\textsc{Pack}$ for hyper-heuristics \\
        $HH_3, HH_4$ & $\textsc{Pack}$ for hyper-heuristics and \\
        ~ & iterative reward functions \\
        $HH_5, HH_6$ & $\textsc{Pack}_{SF}$ for hyper-heuristics and \\
        ~ & iterative reward functions \\
    \hline
\end{tabular}
\end{table}

\subsection{Experimental Results}
\label{sec6_2}
We present the key experimental results for each setting as mean objective scores and standard deviations to compare the effectiveness of existing reward functions with the new ones in both deterministic and stochastic constrained environments.
Table~\ref{table: P1-H-algorithms} presents the results for the PWT with deterministic constraints. It compares the performance of the four new reward functions with the original reward function $r1$. Among the reward functions that calculate the score only once, $r3$ shows better performance across all instances. However, once iterative updates to the rewards are allowed, greedy approaches achieve even better results.
$r5$ shows the best mean values for the majority of the instances. When comparing the new reward functions, the key observation we made is that when a reward function uses the benefit of adding an item (profit minus cost difference) as a heuristic, such as $r2$ and $r4$, it underperforms $r1$. However, the benefit and the weight of the item are brought in as a ratio (benefit-to-weight), which exceeds the performance of the original setting across all instances.

\begin{table*} 
    \caption{Performance of the \textsc{Pack} algorithm with different reward functions on the PWT with deterministic constraints.}
    \label{table: P1-H-algorithms}
    \centering
\resizebox{\textwidth}{!}{%
\begin{tabular}{|c|c c|c c|c c|c c|c c|}
    \hline
    instance & \multicolumn{2}{c}{$r$1} & \multicolumn{2}{|c|}{$r2$} & \multicolumn{2}{|c|}{$r3$} & \multicolumn{2}{|c|}{$r4$} & \multicolumn{2}{|c|}{$r5$} \\
    ~ & mean & std & mean & std & mean & std & mean & std & mean & std \\
    \hline
        unc\_51 & 1368.04 & 476.93 & 1444.64 & 309.07 & \textbf{1820.21} & 169.68 & 1108.05 & 691.93 & 1775.67 & 147.59 \\ 
        unc\_152 & 2754.18 & 921.60 & 2153.71 & 695.75 & \textbf{3727.24} & 1062.50 & 2700.60 & 970.43 & 3701.47 & 1012.03 \\ 
        unc\_280 & 10637.54 & 986.40 & 12683.06 & 661.05 & \textbf{16535.55} & 1531.58 & 12273.59 & 549.02 & 16345.35 & 1322.97 \\ 
        unc\_575 & 20604.12 & 1235.43 & 19020.23 & 2178.85 & 30702.12 & 860.41 & 15587.63 & 1928.82 & \textbf{30833.99} & 631.09 \\ 
        unc\_1000 & 104190.60 & 2489.20 & 90115.00 & 0.00 & \textbf{165486.00} & 0.00 & 90115.00 & 0.00 & \textbf{165486.00} & 0.00 \\ 
        bsc\_51 & 3532.57 & 116.22 & 2806.14 & 170.74 & 3632.57 & 15.52 & 2500.68 & 405.45 & \textbf{3687.53} & 116.94 \\ 
        bsc\_152 & 8038.93 & 195.98 & 4334.62 & 61.28 & 8714.79 & 887.24 & 5200.71 & 70.47 & \textbf{10366.22} & 159.84 \\ 
        bsc\_280 & 13197.18 & 478.54 & 7103.99 & 751.89 & 14127.89 & 1113.72 & 9827.85 & 1555.11 & \textbf{16888.90} & 1290.47 \\ 
        bsc\_575 & 21981.33 & 1621.35 & 2508.19 & 2019.59 & 25226.55 & 1194.84 & 12978.57 & 1077.09 & \textbf{29849.01} & 623.15 \\ 
        bsc\_1000 & 109490.63 & 2117.59 & 99425.00 & 0.00 & \textbf{144219.00} & 0.00 & 99425.00 & 0.00 & \textbf{144219.00} & 0.00 \\
    \hline
\end{tabular}
}
\end{table*}

\begin{table*} 
    \caption{Performance of \textsc{Pack} algorithms with different hyper-heuristics on the PWT with deterministic constraints.}
    \label{table: P1-HH-algorithms}
    \centering
\small
\begin{tabular}{|c|c c|c c|c c|c c|}
    \hline
    instance & \multicolumn{2}{|c|}{$HH_{1}$} & \multicolumn{2}{|c|}{$HH_{2}$} & \multicolumn{2}{|c|}{$HH_{3}$} & \multicolumn{2}{|c|}{$HH_{4}$} \\
    ~ & mean & std & mean & std & mean & std & mean & std \\
    \hline
        unc\_51 & 1864.76 & 171.12 & \textbf{1925.51} & 125.61 & 1876.87 & 232.16 & 1915.51 & 111.23 \\ 
        unc\_152 & 3472.97 & 1120.34 & \textbf{3865.93} & 1078.53 & 3424.79 & 1089.50 & \textbf{3865.93} & 1078.53 \\ 
        unc\_280 & 15734.58 & 671.98 & 16840.14 & 1286.13 & 15590.71 & 946.79 & \textbf{16908.97} & 1265.31 \\ 
        unc\_575 & 29444.28 & 868.48 & 31511.08 & 652.25 & 29227.75 & 1420.34 & \textbf{31767.77} & 606.75 \\ 
        unc\_1000 & 127662.70 & 4200.46 & \textbf{165486.00} & 0.00 & 130883.07 & 4595.76 & \textbf{165486.00} & 0.00 \\ 
        bsc\_51 & 3726.74 & 127.31 & \textbf{3841.73} & 1.94 & 3771.62 & 92.58 & 3838.71 & 8.12 \\ 
        bsc\_152 & 9411.23 & 573.29 & 9897.17 & 429.89 & 9483.09 & 805.12 & \textbf{10526.35} & 260.06 \\ 
        bsc\_280 & 14888.44 & 1150.00 & 16267.19 & 1788.83 & 14981.98 & 1055.24 & \textbf{17078.60} & 1198.00 \\ 
        bsc\_575 & 24794.52 & 1380.44 & 27993.55 & 1234.79 & 24640.65 & 1826.39 & \textbf{29896.53} & 634.26 \\ 
        bsc\_1000 & 114412.13 & 1568.70 & \textbf{144219.00} & 0.00 & 115236.03 & 1615.89 & \textbf{144219.00} & 0.00 \\
    \hline
\end{tabular}

\end{table*}

\begin{table}[t] 
    \caption{Statistical analysis of best performing algorithms for PWT with deterministic constraints.}
    \label{table: stat-test-pwt}
    \centering
\small
\begin{tabular}{|l|l|l|l|l|}
    \hline
   instance & p value & $r5$ & $HH_4$ \\ 
    \hline
        unc\_51 &  0.0002 &  (-) &  (+) \\ 
        unc\_152 &  0.0012 &  (-) &  (+) \\ 
        unc\_280 &  0.0026 &  (-) &  (+) \\ 
        unc\_575 &  1.80e-07 &  (-) &  (+) \\ 
        unc\_1000 &  1.0000 &  * &  * \\ 
        bsc\_51 &  8.26e-11 &  (-) &  (+) \\ 
        bsc\_152 &  0.0004 &  (-) &  (+) \\ 
        bsc\_280 &  0.0345 &  (-) &  (+) \\ 
        bsc\_575 &  0.7901 &  * &  * \\ 
        bsc\_1000 &  1.0000 &  * &  * \\ 
    \hline
\end{tabular}
\end{table}

For the PWT with deterministic constraints, Table~\ref{table: P1-HH-algorithms} presents the performance of the various hyper-heuristic algorithms we tested across different heuristic settings, and compares their outcomes with the original reward function $r1$ coupled with the \textsc{Pack} algorithm.
We first examine the advantage of combining only simple one-time reward functions ($r1$, $r2$ and $r3$) in the hyper-heuristic framework. It provides a significant improvement in the objective scores and reduces standard deviations, compared to the original setting. A further improvement is observed when the LLH combination is initialised using the best-performing individual reward function ($HH_2$). As hyper-heuristic methods are resource-intensive, initialisation is a key aspect for achieving better optimisation within the limited iterations available.
Then, we include the iterative reward functions ($r4$ and $r5$) in the selection and investigate the performance with different initialisation settings. The increased number of LLHs in the mix can lead to diverse combinations of heuristics, which may provide further opportunities to reach better solutions. The results showcase this advantage in all instances. Further, the advantage of using iterative reward functions is more evident when the initialisation is based on the best-performing heuristic.
Overall, the hyper-heuristic approaches tested give higher objective scores for all instances than the original reward function. They perform better when initialisation is based on the best-performing reward function, and when iterative heuristics are involved ($HH_2$ and $HH_4$). $HH_4$ outperforms all other hyper-heuristics settings, except for the smaller instances (unc\_51 and bsc\_51).

Table~\ref{table: P1_SF-algorithms} collectively presents the performances of both the heuristic and hyper-heuristic approaches and compares them against the original reward function $r1$ coupled with the $\textsc{Pack}_{SF}$ algorithm. We use a small uncertainty ($\delta=20$) and two different thresholds for the chance constraint ($\alpha=\{0.9, 0.999\}$), to examine the behaviour of the algorithms. The algorithms utilise either the Hoeffding bound or Chebyshev's inequality based on the threshold in use. When the threshold is tighter, the algorithms find it harder to reach higher optimisation levels, as reflected in the lower mean objective scores for $\alpha=0.999$ in almost all cases.
As shown, we present only the iterative heuristics for stochastic constraints. The reason is that, based on our previous experiments on the deterministic constraint, we realise the iterative heuristics work better than the simple one-time heuristics. Further, since $\textsc{Pack}_{SF}$ needs to calculate the increased expected weights, each time an item is added to the packing plan, it is necessary to recalculate rewards, which naturally makes the reward function iterative.
Similar to the performance observed in the deterministic constraints, the benefit of adding an item used directly as the reward function ($r6$) is not successful over the original reward function in the stochastic setting. The benefit and the increased expected weight combined in a ratio ($r7$) provide a significant improvement.
However, the hyper-heuristic setting $HH_6$, which initialises the heuristic combination with the best-performing reward function, outperforms all other methods across all instances.

\begin{table*} 
    \caption{Performance of $\textsc{Pack}_{SF}$ with different reward functions and hyper-heuristics on PWT with stochastic constraints}
    \label{table: P1_SF-algorithms}
    \centering
\resizebox{\textwidth}{!}{%
\begin{tabular}{|c|c|c|c c|c c|c c|c c|c c|}
    \hline
    instance & $\alpha$ & $\delta$ & \multicolumn{2}{|c|}{$r1$} & \multicolumn{2}{|c|}{$r6$} & \multicolumn{2}{|c|}{$r7$} & \multicolumn{2}{|c|}{$HH_{5}$} & \multicolumn{2}{|c|}{$HH_{6}$}\\ 
    ~ & ~ & ~ & mean & std & mean & std & mean & std & mean & std & mean & std \\
    \hline
        unc\_51 & 0.9 & 20 & 1480.11 & 389.67 & 1108.05 & 680.30 & 1788.41 & 163.12 & 1519.44 & 414.30 & \textbf{1853.82} & 153.64 \\ 
        unc\_51 & 0.999 & 20 & 1451.51 & 409.89 & 1104.94 & 654.95 & 1691.28 & 199.48 & 1464.50 & 428.26 & \textbf{1915.51} & 109.36 \\ 
        unc\_152 & 0.9 & 20 & 2751.33 & 905.66 & 2225.26 & 1238.71 & 3701.47 & 995.02 & 2974.88 & 661.36 & \textbf{3802.90} & 1077.78 \\ 
        unc\_152 & 0.999 & 20 & 2724.08 & 910.39 & 2094.89 & 1312.99 & 3533.24 & 837.81 & 2944.65 & 679.81 & \textbf{3540.27} & 832.94 \\ 
        unc\_280 & 0.9 & 20 & 10297.09 & 983.66 & 11784.74 & 629.92 & 16337.69 & 1293.67 & 14569.46 & 668.68 & \textbf{16634.86} & 1168.34 \\ 
        unc\_280 & 0.999 & 20 & 9806.09 & 1064.92 & 10901.52 & 467.24 & 16285.62 & 1242.00 & 14164.76 & 589.89 & \textbf{16602.22} & 1163.03 \\ 
        unc\_575 & 0.9 & 20 & 20195.44 & 1164.50 & 14827.18 & 1771.62 & 30677.60 & 688.41 & 23318.88 & 1537.12 & \textbf{30745.92} & 711.34 \\ 
        unc\_575 & 0.999 & 20 & 19312.53 & 1227.05 & 14032.89 & 1876.54 & 30337.38 & 664.34 & 22090.88 & 1550.59 & \textbf{30470.06} & 668.71 \\ 
        unc\_1000 & 0.9 & 20 & 102906.63 & 2407.09 & 89150.00 & 0.00 & \textbf{164414.00} & 0.00 & 127484.37 & 1404.88 & \textbf{164414.00} & 0.00 \\ 
        unc\_1000 & 0.999 & 20 & 101787.87 & 2449.01 & 88872.00 & 0.00 & \textbf{163259.00} & 0.00 & 125996.33 & 4374.81 & \textbf{163259.00} & 0.00 \\ 
        bsc\_51 & 0.9 & 20 & 3535.07 & 114.68 & 2842.49 & 567.00 & 3647.34 & 76.10 & 3593.11 & 41.47 & \textbf{3690.34} & 62.49 \\ 
        bsc\_51 & 0.999 & 20 & 3318.30 & 105.25 & 2742.30 & 161.16 & 3484.37 & 21.59 & 3559.28 & 120.85 & \textbf{3648.58} & 25.24 \\ 
        bsc\_152 & 0.9 & 20 & 7843.27 & 116.44 & 5025.25 & 32.39 & 10226.01 & 137.96 & 9299.20 & 791.04 & \textbf{10344.11} & 216.81 \\ 
        bsc\_152 & 0.999 & 20 & 7699.76 & 162.76 & 4827.09 & 171.65 & 10037.06 & 138.93 & 9226.12 & 739.15 & \textbf{10164.30} & 247.06 \\ 
        bsc\_280 & 0.9 & 20 & 13132.68 & 487.96 & 9573.68 & 1583.75 & 16818.31 & 1217.60 & 14920.64 & 889.49 & \textbf{16878.59} & 1217.71 \\ 
        bsc\_280 & 0.999 & 20 & 12915.01 & 471.16 & 9367.55 & 1560.37 & 16754.71 & 1194.45 & 14687.31 & 898.94 & \textbf{16812.78} & 1196.57 \\ 
        bsc\_575 & 0.9 & 20 & 21809.29 & 1531.53 & 12697.51 & 1131.90 & 29689.24 & 560.88 & 23605.91 & 1109.11 & \textbf{29693.90} & 558.57 \\ 
        bsc\_575 & 0.999 & 20 & 21591.31 & 1504.21 & 12732.64 & 1176.07 & 29449.92 & 532.99 & 23521.20 & 1216.99 & \textbf{29454.14} & 532.92 \\ 
        bsc\_1000 & 0.9 & 20 & 108733.47 & 2117.32 & 99033.00 & 0.00 & \textbf{143381.00} & 0.00 & 113512.80 & 1069.06 & \textbf{143381.00} & 0.00 \\ 
        bsc\_1000 & 0.999 & 20 & 108287.53 & 2143.95 & 97840.00 & 0.00 & \textbf{142551.00} & 0.00 & 113011.27 & 1582.90 & \textbf{142551.00} & 0.00 \\
    \hline
\end{tabular}
}

\end{table*}

\begin{table}[t] 
    \caption{Statistical analysis of best performing algorithms for PWT with stochastic constraints.}
    \label{table: stat-test-cc-pwt}
    \centering
\small
\begin{tabular}{|l|l|l|l|l|}
    \hline
   instance & $\alpha$ & p value & $r7$ & $HH_6$ \\ 
    \hline
        unc\_51 & 0.9 & 0.0002 &  (-) &  (+) \\ 
        unc\_51 & 0.999 & 0.0002 &  (-) &  (+) \\ 
        unc\_152 & 0.9 & 0.5027 &  * &  * \\ 
        unc\_152 & 0.999 & 0.0012 &  (-) &  (+) \\ 
        unc\_280 & 0.9 & 0.0493 &  (-) &  (+) \\ 
        unc\_280 & 0.999 & 0.0211 &  (-) &  (+) \\ 
        unc\_575 & 0.9 & 0.3830 &  * &  * \\ 
        unc\_575 & 0.999 & 0.3366 &  * &  * \\ 
        unc\_1000 & 0.9 & 1.0000 &  * &  * \\ 
        unc\_1000 & 0.999 & 1.0000 &  * &  * \\ 
        bsc\_51 & 0.9 & 0.0007 &  (-) &  (+) \\ 
        bsc\_51 & 0.999 & 9.76e-12 &  (-) &  (+) \\ 
        bsc\_152 & 0.9 & 0.0004 &  (-) &  (+) \\ 
        bsc\_152 & 0.999 & 0.0004 &  (-) &  (+) \\ 
        bsc\_280 & 0.9 & 0.3750 &  * &  * \\ 
        bsc\_280 & 0.999 & 0.3750 &  * &  * \\ 
        bsc\_575 & 0.9 & 0.9058 &  * &  * \\ 
        bsc\_575 & 0.999 & 0.8941 &  * &  * \\ 
        bsc\_1000 & 0.9 & 1.0000 &  * &  * \\ 
        bsc\_1000 & 0.999 & 1.0000 &  * &  * \\
    \hline
\end{tabular}
\end{table}

Finally, we conduct statistical tests to assess the significance of the distribution of objective scores of the best-performing algorithms in both constraint settings. In particular, we are interested in whether the use of the hyper-heuristic approach allows for an improvement that is statistically significant over the best single greedy reward function. We use the Kruskal-Wallis test to determine if at least one of the algorithms has a significantly different distribution, and call a result significant if the p-value is less than $0.05$.
Once the statistical significance is verified, we refer to the resulting mean values and denote each algorithm as statistically significantly better or worse using the symbols $(+)$ and $(-)$, respectively.
If the distributions are not significantly different, we denote that using a $*$.
For the results with respect to the deterministic constraints, $HH_4$ significantly outperforms $r5$ in $7$ out of the $10$ instances (see Table~\ref{table: stat-test-pwt}). For the results with respect to the stochastic constraints, $HH_6$ significantly outperforms $r7$ in $5$ of the $10$ considered instances (see Table~\ref{table: stat-test-cc-pwt}). 

\section{Conclusions}
\label{sec7}
Reward functions play a critical role in greedy algorithms as they determine the order in which items are used for selection. In this paper, we introduced a set of tailored reward functions and hyper-heuristic approaches for a well-known greedy algorithm, namely \textsc{Pack}, to solve PWT.
Our approach captures the exact benefit of having an item in the packing plan and utilises that insight within the reward function to influence the packing order. Through extensive experiments, we investigated the impact of new methods relative to the original reward function, under both deterministic and stochastic constraints. Overall, our tailored reward functions based on the benefit-to-weight ratio show a significant improvement in the packing optimisation across all constraint settings. The hyper-heuristic approaches revealed additional improvements achievable by combining new reward functions at an additional cost.

\section*{Acknowledgement}
This work was supported by the Australian Research Council (ARC) through grant FT200100536 and by supercomputing resources provided by the $Phoenix$ HPC service at Adelaide University.

\bibliographystyle{unsrt}
\bibliography{main}

@inproceedings{pathiragedon2025weighted,
    author = {Pathirage Don, Thilina and Neumann, Aneta and Neumann, Frank},
    title = {Weighted-Scenario Optimisation for the Chance Constrained Travelling Thief Problem},
    booktitle = {IEEE Congress on Evolutionary Computation},
    publisher = {IEEE},
    pages = {1-8},
    year = {2025}
}

@inproceedings{ross2003learning,
    author = {Peter Ross and Javier G. Mar{\'{\i}}n{-}Bl{\'{a}}zquez and Sonia Schulenburg and Emma Hart},
    title = {Learning a Procedure That Can Solve Hard Bin-Packing Problems:{A} New GA-Based Approach to Hyper-heuristics},
    booktitle = {Genetic and Evolutionary Computation Conference. Proceedings, Part {II}},
    series = {Lecture Notes in Computer Science},
    volume = {2724},
    pages = {1295--1306},
    publisher = {Springer},
    year = {2003}
}

@article{dijkstra1959note,
  title = {A note on two problems in connexion with graphs},
  author = {Dijkstra, Edsger W},
  journal = {Numerische mathematik},
  volume = {1},
  number = {1},
  pages = {269--271},
  year = {1959},
  publisher = {Springer}
}

@article{kruskal1956shortest,
  title={On the shortest spanning subtree of a graph and the traveling salesman problem},
  author={Kruskal, Joseph B},
  journal={Proceedings of the American Mathematical society},
  volume={7},
  number={1},
  pages={48--50},
  year={1956},
  publisher={JSTOR}
}

@incollection{krause2014submodular,
  author       = {Andreas Krause and
                  Daniel Golovin},
  title        = {Submodular Function Maximization},
  booktitle    = {Tractability},
  pages        = {71--104},
  publisher    = {Cambridge University Press},
  year         = {2014}
}

@inproceedings{pathiragedon2025evolutionary,
    author = {Pathirage Don, Thilina and Neumann, Aneta and Neumann, Frank},
    title = {Evolutionary Multitasking for the Scenario-based Travelling Thief Problem},
    booktitle = {Proceedings of the Genetic and Evolutionary Computation Conference},
    publisher = {ACM},
    pages = {809–817},
    year = {2025}
}

@article{polyakovskiy2017packing,
  author = {Sergey Polyakovskiy and Frank Neumann},
  title = {The Packing While Traveling Problem},
  journal = {Eur. J. Oper. Res.},
  volume = {258},
  number = {2},
  pages = {424--439},
  year = {2017}
}

@inproceedings{gao2018randomized,
    author = {Gao, Wanru and Friedrich, Tobias and Neumann, Frank and Hercher, Christian},
    title = {Randomized greedy algorithms for covering problems},
    booktitle = {Proceedings of the Genetic and Evolutionary Computation Conference},
    publisher = {ACM},
    pages = {309–315},
    year = {2018}
}

@inproceedings{martins2017heuristic,
    author = {Martins, Marcella S. R. and El Yafrani, Mohamed and Delgado, Myriam R. B. S. and Wagner, Markus and Ahiod, Bela\"{\i}d and L\"{u}ders, Ricardo},
    title = {HSEDA: a heuristic selection approach based on estimation of distribution algorithm for the travelling thief problem},
    booktitle = {Proceedings of the Genetic and Evolutionary Computation Conference},
    publisher = {ACM},
    pages = {361–368},
    year = {2017}
}

@article{yafrani2018hyperheuristic,
    author = {El Yafrani, Mohamed and Martins, Marcella and Wagner, Markus and Ahiod, Bela\"{\i}d and Delgado, Myriam and L\"{u}ders, Ricardo},
    title = {A hyperheuristic approach based on low-level heuristics for the travelling thief problem},
    journal = {Genetic Programming and Evolvable Machines},
    volume = {19},
    number = {1–2},
    publisher = {Kluwer Academic Publishers},
    pages = {121–150},
    year = {2018}
}

@inproceedings{ali2021hyperheuristic,
    author={Ali, Fathelrahman and Mohamedkhair, Mohamedelfatih},
    title={Hyper-Heuristic Approaches for the Travelling Thief Problem}, 
    booktitle={International Conference on Computer, Control, Electrical, and Electronics Engineering}, 
    publisher = {IEEE},
    pages={1-6},
    year={2021}
}

@article{rodriguez2022sequence,
  title={A sequence-based hyper-heuristic for traveling thieves},
  author={Rodr{\'\i}guez, Daniel and Cruz-Duarte, Jorge M and Ortiz-Bayliss, Jos{\'e} Carlos and Amaya, Ivan},
  journal={Applied Sciences},
  volume={13},
  number={1},
  publisher={MDPI},
  pages={56},
  year={2022}
}

@article{sarkar2024travelling,
    author = {Tamalika Sarkar and Chandrasekharan Rajendran},
    title = {Travelling thief problem: a survey of recent variants, solution approaches and future directions},
    journal = {International Journal of Systems Science: Operations \& Logistics},
    volume = {11},
    number = {1},
    publisher = {Taylor \& Francis},
    pages = {2424200},
    year = {2024}
}

@inproceedings{bonyadi2013travelling,
    author = {Mohammad Reza Bonyadi and Zbigniew Michalewicz and Luigi Barone},
    title = {The travelling thief problem: The first step in the transition from theoretical problems to realistic problems},
    booktitle = {2013 IEEE Congress on Evolutionary Computation},
    publisher = {IEEE},
    pages = {1037--1044},
    year = {2013}
}

@inproceedings{roostapour2019analysis,
    author = {Roostapour, Vahid and Pourhassan, Mojgan and Neumann, Frank},
    title = {Analysis of baseline evolutionary algorithms for the packing while travelling problem},
    booktitle = {Proceedings of the 15th ACM/SIGEVO Conference on Foundations of Genetic Algorithms},
    publisher = {ACM},
    pages = {124–132},
    year = {2019}
}

@article{neumann2025optimizing,
    author = {Neumann, Aneta and Neumann, Frank},
    title = {Optimizing monotone chance-constrained submodular functions using evolutionary multiobjective algorithms},
    journal = {Evolutionary Computation},
    volume = {33},
    number = {3},
    publisher = {MIT Press},
    pages = {363--393},
    year = {2025}
}

@article{miller1965chance,
    author = {Bruce L Miller and Harvey M Wagner},
    title = {Chance Constrained Programming with Joint Constraints},
    journal = {Operations Research},
    volume = {13},
    issue = {6},
    pages = {930--945},
    year = {1965}
}

@inproceedings{kohout1999time,
    author = {Kohout, Robert and Erol, Kutluhan and Robert, C},
    title = {In-time agent-based vehicle routing with a stochastic improvement heuristic},
    booktitle = {Proceedings of the 16th National Conference on Artificial Intelligence},
    publisher = {AAAI},
    pages = {864--869},
    year = {1999}
}

@article{applegate2003chained,
    author = {David Applegate and Wiliam Cook and Andre Rohe},
    title = {Chained Lin-Kernighan for large traveling salesman problems},
    journal = {INFORMS Journal on Computing; Winter},
    volume = {15},
    issue = {1},
    pages = {82--92},
    year = {2003}
}

@inproceedings{bent2003dynamic,
    author = {Bent, Russell and Van Hentenryck, Pascal},
    title = {Dynamic vehicle routing with stochastic requests},
    booktitle = {Proceedings of the 18th International Joint Conference on Artificial Intelligence},
    publisher = {Morgan Kaufmann},
    pages = {1362--1363},
    year = {2003}
}

@inproceedings{bent2007waiting,
    title = {Waiting and Relocation Strategies in Online Stochastic Vehicle Routing},
    author = {Russell Bent and Pascal Van Hentenryck},
    booktitle = {Proceedings of the 20th International Joint Conference on Artificial Intelligence},
    publisher = {Morgan Kaufmann},
    pages = {1816--1821},
    year = {2007}
}

@inproceedings{bonyadi2014socially,
    author = {Mohammad Reza Bonyadi and Zbigniew Michalewicz and Michał Roman Przybyłek and Adam Wierzbicki},
    title = {Socially inspired algorithms for the traveling thief problem},   
    booktitle = {Proceedings of the Genetic and Evolutionary Computation Conference},
    publisher = {ACM},    
    pages = {421--428},
    year = {2014}
}

@inproceedings{faulkner2015approximate,
    author = {Hayden Faulkner and Sergey Polyakovskiy and Tom Schultz and Markus Wagner},
    title = {Approximate approaches to the traveling thief problem},
    booktitle = {Proceedings of the Genetic and Evolutionary Computation Conference},
    publisher = {ACM},   
    pages = {385--392},
    year = {2015}
}

@inproceedings{yafrani2016population,
    author = {Mohamed El Yafrani and Belaïd Ahiod},
    title = {Population-based vs. Single-solution heuristics for the Travelling Thief Problem},
    booktitle = {Proceedings of the Genetic and Evolutionary Computation Conference},
    publisher = {ACM},    
    pages = {317--324},
    year = {2016}
}

@inproceedings{wu2017exact,
    author = {Junhua Wu and Markus Wagner and Sergey Polyakovskiy and Frank Neumann},
    title = {Exact Approaches for the Travelling Thief Problem},
    booktitle = {Proceedings of the Simulated Evolution and Learning - 11th International Conference},
    series = {Lecture Notes in Computer Science},
    volume = {10593},
    publisher = {Springer},
    pages = {110--121},
    year = {2017}
}

@article{yafrani2018efficiently,
    author = {Mohamed El Yafrani and Belaïd Ahiod},
    title = {Efficiently solving the Traveling Thief Problem using hill climbing and simulated annealing},
    journal = {Information Sciences},
    volume = {432},
    pages = {231--244},
    year = {2018}
}

@inproceedings{wuijts2019investigation,
    author = {Rogier Hans Wuijts and Dirk Thierens},
    title = {Investigation of the traveling thief problem},
    booktitle = {Proceedings of the Genetic and Evolutionary Computation Conference},
    publisher = {ACM},
    pages = {329--337},
    year = {2019}
}

@inproceedings{xie2019evolutionary,
    author = {Yue Xie and Oscar Harper and Hirad Assimi and Aneta Neumann and Frank Neumann},
    title = {Evolutionary algorithms for the chance-constrained knapsack problem},
    booktitle = {Proceedings of the Genetic and Evolutionary Computation Conference},
    publisher = {ACM},
    pages = {338--346},
    year = {2019}
}

@inproceedings{namazi2019profit,
    author = {Majid Namazi and M. A. Hakim Newton and Abdul Sattar and Conrad Sanderson},
    title = {A Profit Guided Coordination Heuristic for Travelling Thief Problems},
    booktitle = {Proceedings of the 12th International Symposium on Combinatorial Search},
    publisher = {{AAAI} Press},
    pages = {140--144},
    year = {2019}
}

@inproceedings{doerr2020optimization,
    author = {Benjamin Doerr and Carola Doerr and Aneta Neumann and Frank Neumann and Andrew M. Sutton},
    title = {Optimization of Chance-Constrained Submodular Functions},
    booktitle = {Proceedings of the 34th {AAAI} Conference on Artificial Intelligence},
    publisher = {{AAAI} Press},
    pages = {1460--1467},
    year = {2020}
}

@article{maity2020efficient,
    author = {Alenrex Maity and Swagatam Das},
    title = {Efficient hybrid local search heuristics for solving the travelling thief problem},
    journal = {Appl. Soft Comput.},
    volume = {93},
    pages = {106284},
    year = {2020}
}

@inproceedings{shi2022runtime,
    author = {Feng Shi and Xiankun Yan and Frank Neumann},
    title = {Runtime Analysis of Simple Evolutionary Algorithms for the Chance-Constrained Makespan Scheduling Problem},
    booktitle = {Parallel Problem Solving from Nature - 17th International Conference, Proceedings, Part {II}},
    series = {Lecture Notes in Computer Science},
    volume = {13399},
    publisher = {Springer},
    pages = {526--541},
    year = {2022}
}

@inproceedings{yan2023optimizing,
    author = {Xiankun Yan and Anh Viet Do and Feng Shi and Xiaoyu Qin and Frank Neumann},
    title = {Optimizing Chance-Constrained Submodular Problems with Variable Uncertainties},
    booktitle = {Proceedings of the 26th European Conference on Artificial Intelligence},
    series = {Frontiers in Artificial Intelligence and Applications},
    volume = {372},
    publisher = {{IOS} Press},    
    pages = {2826--2833},
    year = {2023}
}

@inproceedings{polyakovskiy2014comprehensive,
    author={Polyakovskiy, Sergey and Bonyadi, Mohammad Reza and Wagner, Markus and Michalewicz, Zbigniew and Neumann, Frank},
    title={A comprehensive benchmark set and heuristics for the traveling thief problem},
    booktitle={Proceedings of the Genetic and Evolutionary Computation Conference},
    publisher = {ACM},
    pages={477--484},
    year={2014}
}

@inproceedings{yan2024sampling,
    author = {Xiankun Yan and Aneta Neumann and Frank Neumann},
    title = {Sampling-based Pareto optimization for chance-constrained monotone submodular problems},
    booktitle = {Proceedings of the Genetic and Evolutionary Computation Conference},
    publisher = {ACM},
    pages = {621--629},
    year = {2024}
  }

@inproceedings{neumann2021diversifying,
    author = {Aneta Neumann and Jakob Bossek and Frank Neumann},
    title = {Diversifying greedy sampling and evolutionary diversity optimisation for constrained monotone submodular functions},
    booktitle = {Proceedings of the Genetic and Evolutionary Computation Conference},
    publisher = {ACM},
    pages = {261--269},
    year = {2021}
}

@inproceedings{neumann2022evolutionary,
    author = {Aneta Neumann and Yue Xie and Frank Neumann},
    title = {Evolutionary algorithms for limiting the effect of uncertainty for the knapsack problem with stochastic profits},
    booktitle = {Parallel Problem Solving from Nature - 17th International Conference, Proceedings, Part {I}},
    series = {Lecture Notes in Computer Science},
    volume = {13398},
    publisher = {Springer},
    pages = {294--307},
    year = {2022}
}

@article{li2024chance,
    author = {Li, Xuanfeng and Liu, Shengcai and Wang, Jin and Chen, Xiao and Ong, Yew-Soon and Tang, Ke}, 
    title = {Chance-Constrained Multiple-Choice Knapsack Problem: Model, Algorithms, and Applications},
    journal = {{IEEE} Trans. Cybern.},
    volume = {54},
    number = {12},
    pages = {7969--7980},
    year = {2024}
}

@inproceedings{pathiragedon2024thechance,
    author = {Pathirage Don, Thilina and Neumann, Aneta and Neumann, Frank},
    title = {The Chance Constrained Travelling Thief Problem: Problem Formulations and Algorithms},
    booktitle = {Proceedings of the Genetic and Evolutionary Computation Conference},
    publisher = {ACM},
    pages = {214--222},
    year = {2024}
}

@article{zhang2021solving,
    author={Zhang, Zitong and Yang, Lei and Kang, Peipei and Jia, Xiaotian and Zhang, Wensheng},
    journal= {IEEE Access}, 
    title = {Solving the Traveling Thief Problem Based on Item Selection Weight and Reverse-Order Allocation}, 
    year = {2021},
    volume = {9},
    number = {},
    pages = {54056-54066}
}
\end{document}